# ETRI-Activity3D: A Large-Scale RGB-D Dataset for Robots to Recognize Daily Activities of the Elderly

Jinhyeok Jang, Dohyung Kim*, Cheonshu Park, Minsu Jang, Jaeyeon Lee, Jaehong Kim

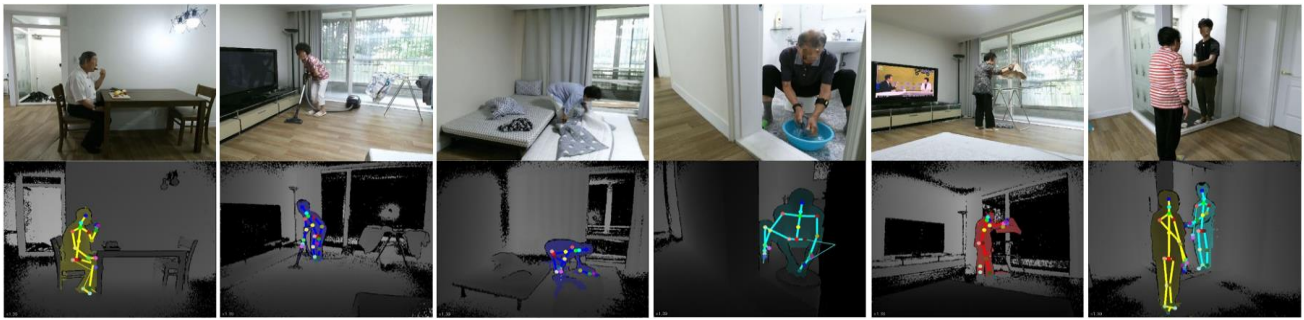

Figure 1. Examples of daily actions in the proposed dataset are displayed along with the corresponding depth map and skeleton information obtained from the Kinect v2 sensors. Actions (from left to right): *eating food with a fork, vacuuming the floor, spreading the bedding, washing a towel by hands, hanging out laundry, and hand shaking*. The entire dataset is available for download at the following link: https://ai4robot.github.io/etri-activity3d-en.

*Abstract—* Deep learning, based on which many modern algorithms operate, is well known to be data-hungry. In particular, the datasets appropriate for the intended application are difficult to obtain. To cope with this situation, we introduce a new dataset called ETRI-Activity3D, focusing on the daily activities of the elderly in robot-view. The major characteristics of the new dataset are as follows: 1) practical action categories that are selected from the close observation of the daily lives of the elderly; 2) realistic data collection, which reflects the robot's working environment and service situations; and 3) a large-scale dataset that overcomes the limitations of the current 3D activity analysis benchmark datasets. The proposed dataset contains 112,620 samples including RGB videos, depth maps, and skeleton sequences. During the data acquisition, 100 subjects were asked to perform 55 daily activities. Additionally, we propose a novel network called four-stream adaptive CNN (FSA-CNN). The proposed FSA-CNN has three main properties: robustness to spatio-temporal variations, input-adaptive activation function, and extension of the conventional two-stream approach. In the experiment section, we confirmed the superiority of the proposed FSA-CNN using NTU RGB+D and ETRI-Activity3D. Further, the domain difference between both groups of age was verified experimentally. Finally, the extension of FSA-CNN to deal with the multimodal data was investigated.

## I. INTRODUCTION

The tremendous success of deep learning approaches has brought a substantial improvement in several computer vision tasks. Because these approaches are heavily dependent on large and reliable datasets, efforts have been made to create such datasets, resulting in several publicly available datasets. This is also the case with human action understanding [1, 2, 9, 10, 11]. However, the characteristics of the reported datasets are biased toward their intended applications.

In this study, we employ the recognition of daily activities from a robot's point of view. We believe that the elderly in particular would be the first serious users of robot services. Therefore, they are primarily assumed to be the users. However, evidently, the reported datasets do not fit in this important application.

Therefore, ETRI-Activity3D, which precisely targets the application of recognition of daily activities of the elderly from the robot's point of view, is employed. The unique characteristics of this proposed dataset over the existing ones are provided in the following.

**A new visual dataset based on observations of the daily activities of the elderly:** A close understanding of what the elderly actually do in their daily lives is essential for constructing a useful dataset. Therefore, we visited the homes of 53 elderly people over the age of 70 years and carefully monitored and documented their daily behavior from morning to night. Then, we selected 55 most frequent actions. Further, while constructing the dataset, we recruited 50 elderly people aged between 64 and 88 years, which led to a realistic intra-class variation of the actions. Additionally, we recruited 50 young people in their 20s. This composition allowed us to perform various comparative experiments between the behavior of the elderly and that of the young. Thus, we were provided with a deeper understanding of the behavioral characteristics of the elderly.

**A realistic dataset considering the service situation of human-care robots:** The aim of the proposed dataset is to be utilized in practical research that can be applied to real-world environments. Therefore, while designing the dataset, probable situations that can occur when robots are in service are closely investigated. The data acquisition is performed in an apartment that has conditions quite similar to the living conditions of the elderly. Each subject is advised to perform the given action in his/her own way. Additionally, the subjects' actions are carried out in diverse environmental conditions such as the time of the day or places in the apartment and in various human postures. We also acquired data from the probable locations where the robot is supposed to be when it serves humans. For instance, in small spaces such as kitchens or bathrooms, the proposed dataset includes scenes where humans face their backs while performing the actions. Although recognizing the actions from their backs is difficult, we believe that these actions are realistic situations that may occur frequently. These considerations make the proposed dataset more challenging, but they make the dataset more practical.

**A large-scale RGB-D action recognition dataset that overcomes the limitations of the reported datasets:** A sufficient amount of data is crucial for developing deep learning algorithms. However, building a sufficiently large dataset is extremely expensive and difficult, especially for 3D actions, because there is no proper way to leverage video-sharing services or crowdsourcing as in their 2D counterparts. Because of these difficulties, only a limited number of 3D datasets have been reported in several aspects. Among them, the major drawbacks are a small number of subjects and action categories and monotonous environmental conditions. As presented in Table I, the proposed dataset consists of 112,620 video samples, which is comparable to the current largest 3D action dataset, NTU RGB+D 120 [10]. A dataset of this scale with realistic variations may aid in researching extensively from a variety of perspectives, which are expected to contribute to the advancement of robotics intelligence research.

Additionally, we propose a novel action recognition network, four-stream adaptive CNN (FSA-CNN). This proposed network was used to include the spatio-temporal variation of action data. FSA-CNN has three main properties: global pooling that provides the robustness to temporal variations, activation network that helps in input-specific adaptation, and four-stream inputs. Each property is explained in detail in section IV. The proposed network can recognize the actions of not only the elderly, but also the adults well. We evaluated the proposed FSA-CNN using the NTU RGB+D and ETRI-Activity3D datasets. We then analyzed the domain differences and robustness to temporal variations.

In summary, the major contributions of this study are as follows: 1) a new large-scale dataset for the action recognition of the elderly is introduced, 2) a novel neural architecture that is robust to realistic variations is proposed, and 3) the efficiency and usefulness of the dataset are validated using extensive experiments.

TABLE I. COMPARISON BETWEEN ETRI-ACTIVITY3D AND OTHER PUBLICLY AVAILABLE DATASETS FOR 3D DAILY ACTIVITY RECOGNITION. DATA MODALITIES: (RGB)VIDEO, (D)EPTH, (S)KELETON, (I)R.

| Datasets | #Samples | #Sub | #Act | Modalities |
|---|---|---|---|---|
| RGBD-HuDaAct [3] | 1,189 | 30 | 13 | RGBD |
| MSRDailyActivity3D [4] | 320 | 10 | 16 | RGBDS |
| Act4$^2$ [5] | 6,844 | 24 | 14 | RGBD |
| CAD-120 [6] | 120 | 4 | 10+10 | RGBDS |
| Office Activity [7] | 1,180 | 10 | 20 | RGBD |
| UWA3D Multiview II [8] | 1,075 | 10 | 30 | RGBDS |
| NTU RGB+D [9] | 56,880 | 40 | 60 | RGBDSI |
| NTU RGB+D 120 [10] | 114,480 | 106 | 120 | RGBDSI |
| Toyota Smarthome [11] | 16,129 | 18 | 31 | RGBDS |
| **ETRI-Activity3D** | **112,620** | **100** | **55** | **RGBDS** |

## II. RELATED WORK

In this section, we briefly review the publicly available datasets for 3D daily activity recognition and the recent deep learning methods for human action recognition.

### A. 3D daily activity recognition datasets

Table I shows the most popular public datasets captured indoors using the Kinect sensors. Although each one has its own unique characteristics, it has limitations.

RGBD-HuDaAct [3] is one of the largest datasets for the home-monitoring-oriented activity recognition. It contains 1,189 synchronized color-depth video streams that provide rich intra-class variations. However, because the background is limited to a single lab environment with a fixed camera, the dataset is not suitable for practical benchmarks.

MSRDailyActivity3D [4] is designed to include the human daily activities in the living room. It contains 320 samples of 16 activities performed by 10 actors, either sitting on the sofa or standing close to it. However, the small number of samples and fixed camera viewpoints are the limitations of this dataset.

CAD-120 [6] focuses on high-level activities and object interactions. It comprises 120 long-term activities, such as *making cereal* and *microwaving food*. Each video is annotated with human skeleton tracks, object tracks, object affordance labels, sub-activity labels, and high-level activities.

Toyota Smarthome [11] is a real-world video dataset for activities of daily living. It consists of 16,129 RGB+D clips of 31 activity classes performed by the elderly in a smart home. Unlike the other datasets, the videos were completely unscripted in this dataset; this brought out several real-world challenges. However, the limited number of subjects and activity classes are the drawbacks of this dataset.

NTU-RGB+D 120 [10], an extension of NTU-RGB+D [9], is the current largest benchmark dataset for 3D action recognition. It includes 114,480 video samples collected from 120 action classes in multi-view settings. This large-scale dataset has contributed to the development and evaluation of data-driven learning methods. However, this dataset was acquired in a laboratory environment, and the activities were performed with a strict guidance; these do not reflect the realistic challenges that exist in daily activities.

Here, we summarize the main characteristics of ETRI-Activity3D: (1) the daily activities of the elderly were recorded from the robot's point of view; (2) action classes were determined based on observations of the daily activities of the elderly; (3) the dataset acquisition was performed in an apartment reflecting the living conditions of the elderly, and not in a laboratory; (4) the probable service situation of the human-care robots was considered; (5) this dataset provides large variations in views, distances, and backgrounds; and (6) this dataset is the second largest dataset in the daily living activity recognition domain in terms of the number of subjects and video samples.

*B. Vision-based action recognition*

Three types of visual information are mainly used for recognizing action: RGB, depth, and skeleton. Most of the studies recognize action using one or more of these data.

[12, 13] extracted the features from RGB frames. Glimpse Clouds [12] extracts local features from video frames based on an attention model. In the I3D [13] case, the features were extracted from RGB frames and their flows.

Several other studies [14–21] use skeletal data. Skeletal data and their differential are converted into two maps. They are used as inputs in SK-CNN [14], HCN [15], and Ensem-NN [16]. Beyond Joint [17] and IndRNN [18] attempted to use an RNN for action recognition. Beyond Joint proposed a network composed of multiple bi-directional LSTMs, and IndRNN proposed their own layer, Independently RNN (IndRNN), which is easy to train and stack. MANs [19] is composed of a GRU and ResNet-like network that extract temporal and spatial features, respectively. ST-GCN [20] and Motif ST-GCN [21] applied a graph convolutional network (GCN) to handle the skeletal data. These data were represented as joints and bones, and they could be used as vertices and edges of the graph. This concept is suitable for skeleton-based action recognition.

Additionally, few studies combine the multiple sources of data. c-ConvNet [22] combines the features extracted from forward/backward differential maps of both depth and RGB. Then, the pose estimation map [23], which constructs two input maps from an RGB video and a skeleton sequence based on the connectivity of joints, is generated. Deep bilinear learning [24] predicts the actions by combining the features extracted from RGB, depth, and skeletal data using their own bilinear blocks.

Our approach is based on the two-stream method [25, 26]. We extended the two-stream method to four-stream to extract robust features based on the long-term temporal and spatial information. We used activation networks instead of activation layers. Our work has two advantages: robustness to spatio-temporal variation of skeletal data and adaptability to environmental conditions. Additionally, we applied the 2D skeleton data estimated from the RGB video into the proposed FSA-CNN, combined the features extracted from both the 3D and 2D skeletons, and investigated them to achieve better performance.

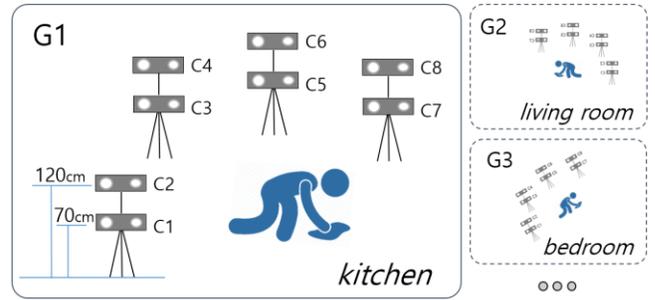

Figure 2. Configuration of the data acquisition system.

TABLE II. ANALYSIS OF THE DIFFERENCES BETWEEN ACTIONS PERFORMED BY THE ELDERLY AND ADULTS OBSERVED ON THE ETRI-ACTIVITY3D DATASET.

|  | Elderly | Adult |
|---|---|---|
| Avg. Frame Length | 267 | 189 |
| Var. Frame Length | 1.189e+04 | 5.663e+03 |
| Avg. Motion Differential | 16.79 | 20.29 |
| Var. Motion Differential | 1.519e+05 | 4.458e+05 |

## III. THE ETRI-ACTIVITY3D DATASET

The ETRI-Activity3D dataset is collected using Kinect v2 sensors, and it consists of three synchronized data modalities: RGB video, depth map, and skeleton sequence. The resolution of RGB video is $1920 \times 1080$, and the depth map is stored frame by frame in a $512 \times 424$ resolution. The skeleton sequence contains the 3D locations of 25 body joints of tracked human bodies. There are 55 action categories, of which 52 are derived from the observation of daily activities (eating, cleaning, reading, etc.) of the elderly and the rest 3 are human-robot interaction specific actions (waving, beckoning, and pointing). Among them, there are five mutual actions such as handshaking and hugging.

The number of subjects was 100, of which 50 were senior citizens, and the rest were young adults. The age of the elderly subjects ranged from 64–88 years with an average of 77 years, whereas the young subjects were in their 20s with an average of 23 years. Among the elderly, 17 were men and 33 were women, whereas the numbers of men and women were 25 for young adults.

When collecting the data, the expected robot view was considered. That is, the capturing device was located at heights of 70 cm and 120 cm, which are based on the typical height of human-care robots, as shown in Fig. 2. Four capturing platforms (each one capturing from both the aforementioned heights) were arranged to capture various views of the action simultaneously. Additionally, the distance between the sensor and subject varied from 1.5–3.5 m. The actions that could be carried out independent of places (e.g., taking medicine or talking on the phone) were captured up to five times at all different places. In this way, we could provide additional intra-class variations in views and backgrounds.

The subjects were advised to ignore the cameras to capture actions as natural as possible. Different postures were recommended while carrying out the action (e.g., taking medicine while sitting or standing, etc.). Additionally, different shapes of relevant objects were provided to the subjects, and they were asked to hold the objects with their right or left hand.

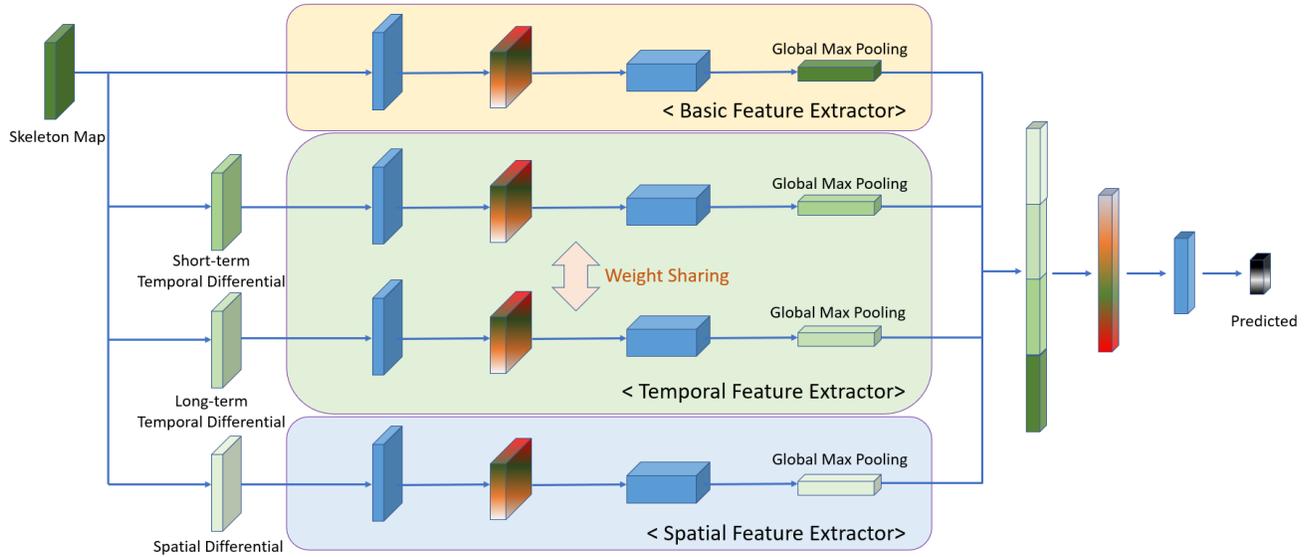

Figure 3. Schematic of the proposed architecture for action recognition.

Thus, we collected a large-scale RGB-D dataset with 112,620 samples. Detailed information on the dataset can be found at the following link: https://ai4robot.github.io/etri-activity3d-en.

Table II shows the differences between actions performed by the elderly and adults observed on the ETRI-Activity3D dataset. The first two rows clearly indicate that the elderly act slower than the adult. The frame length and motion differentials were calculated using normalized skeletons, and three action classes were excluded because of the strong noise. This statistical analysis indicates that the elderly act quite differently from the young, which suggests that the elderly subjects should be included in building realistic datasets.

## IV. FOUR-STREAM ADAPTIVE CNN

The overall network of FSA-CNN is illustrated in Fig. 3. The proposed FSA-CNN has three major properties: robustness to temporal variations, activation network [27], and four-stream inputs. Each row in the figure represents the four streams, and the gradually colored layers represent the activation networks. Each stream has a global max pooling layer at the end to accommodate the variable length inputs. In this section, each component of the proposed FSA-CNN is described. The code can be found at the following link: https://github.com/ai4r/AIR-Action- Recognition.

### A. Architecture with indefinite input-length

The first property, robustness to temporal variations, needs to be addressed because of the presence of innate variations in action lengths, inevitable noises caused by the imperfect detection of action, or domain differences shown in Table II. This property has also been briefly addressed in [20].

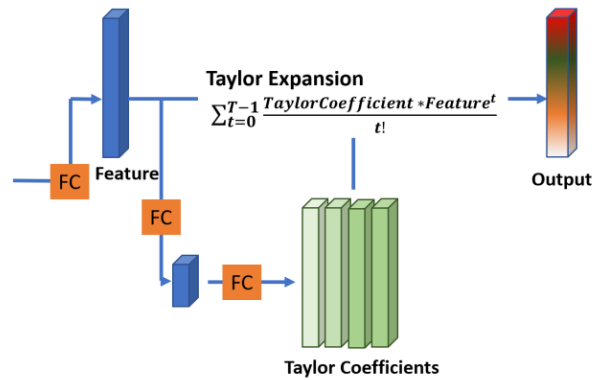

Figure 4. Schematic of a layer in the activation network.

However, several action recognition approaches [12, 14, 15, 24, 25, 26] simply adopt a fixed video length by sampling, padding, and/or interpolating the input video sequence. These modifications often result in information loss.

Therefore, we propose a novel network architecture that is robust against the temporal variations using global max pooling. By adopting this layer, the proposed network can extract robust features with a fixed length even for a varying input video length.

The problems caused by these varying lengths should be considered during the training phase. Therefore, in the training phase, we randomly sampled the data in different lengths and trained the network using the data. This approach has the effect of data augmentation because randomly sampling into various lengths has much more degree of freedom than sampling into a fixed length.

## B. Activation network

The activation network [27] is a substitute for the activation layer. It introduces non-linearity into deep neural networks. The activation layer has a fixed non-linear function such as ReLU, and a deep neural network formulates a high-order non-linearity using multiple activation layers. In contrast, the activation network is a trainable layer that can fit complex curves well without employing multiple layers.

This activation network has two advantages. First, it can reduce the number of trainable weights in a network. In case of the conventional activation layer, multiple layers have to be stacked to fit high-order curves, and a huge amount of data is required to fit such a high-order function well. On the contrary, the activation network can model a higher-order function using only a few trainable weights. Second, the activation curves generated by the activation network are functions of the input data itself. That is, each input has its own activation function. This input-dependent activation function adapts better to input data and thus more robust to data variations when compared to the conventional activation layer. Fig. 4 illustrates the structure of activation network. For a layer, the output of the $i^{th}$ node represented as $u_i^l$ is obtained by the following equation:

$$u_i^l = \sum_j^{n_{l-1}} W_{ij}^l x_j^{l-1} + b_i^l, \qquad (1)$$

for $i = 1, 2, \cdots, n_l$; where $x_j^{l-1}$ is the output of the $j^{th}$ node obtained from the $(l-1)$th layer; and $W_{ij}^l$ and $b_i^l$ are the weight and bias, respectively. A set of polynomial coefficients is obtained by a branch network expressed as per the following equation:

$$a_{ki}^l = \sum_j^{n_l} V_{jk}^l u_j^l + z_{ki}^l, \qquad (2)$$

for k = 0, 1, $\cdots$, K; where $V_{jk}^l$ and $z_{ki}^l$ are the weight and bias for the k$^{th}$-order coefficient of the $i^{th}$ node, $a_{ki}^l$, respectively. The intermediate output is activated by a K$^{th}$-order polynomial function as shown in the following:

$$x_i^l = \sum_k^K a_{ki}^l (u_j^l)^k, \qquad (3)$$

The activation layer generates a set of Taylor coefficients and creates a feature using the weighted sum of Taylor coefficients and numerical powers of the input feature. Thus, a convolution operation can be re-written as a matrix multiplication using the Toeplitz matrix, so that it can be easily applied to the convolution layer.

## C. Four-stream input

The two-stream approach is widely used for action recognition [14, 15, 16, 25, 26]. Several reported action recognition studies adopt two-stream networks that utilize both action and temporal differential sequences as inputs. We extended this concept to the following four streams: action, short-term temporal differential, long-term temporal differential, and spatial differential sequences. Basically, we transformed a skeleton sequence into evolution images as suggested in [23] and fed them into the first stream of Fig. 3. The short-term and long-term temporal differential sequences are differentials of skeleton sequences in the temporal domain. The short-term one is differentiated by a small differential gap, whereas the long-term one is differentiated by a larger gap. Because both the differentials are the types of skeletal motions, a shared network can be used to extract features.

TABLE III. PERFORMANCES OF THE PROPOSED AND STATE-OF-THE-ART METHODS ON NTU RGB+D AND ETRI-ACTIVITY3D. CS: CROSS-SUBJECT, CV: CROSS-VIEW.

| Method | NTU RGB+D | | ETRI-Activity3D |
|---|---|---|---|
| | CS (%) | CV (%) | CS (%) |
| IndRNN [18] | 81.8 | 88.0 | 73.9 |
| Beyond Joint [17] | 79.5 | 87.6 | 79.1 |
| SK-CNN [14] | 83.2 | 89.3 | 83.6 |
| ST-GCN [20] | 81.5 | 88.3 | 86.8 |
| Motif ST-GCN [21] | 84.2 | 90.2 | 89.9 |
| Ensem-NN [16] | 85.1 | 91.3 | 83.0 |
| MANs [19] | 83.0 | 90.7 | 82.4 |
| HCN [15] | 86.5 | 91.1 | 88.0 |
| **FSA-CNN** | **88.1** | **92.2** | **90.6** |

The spatial differential sequence is a differential between two spatially adjacent joints. Because an action contains both the spatial and temporal information, we needed to add a spatial differential sequence in addition to the traditional two-stream method.

## V. EXPERIMENTS

We evaluated the proposed architecture by comparing it with the other state-of-the-art methods using ETRI-Activity3D and NTU RGB+D [9] datasets. Additionally, we analyzed the ETRI-Activity3D using the results of the proposed FSA-CNN to reveal the differences between the data of the elderly and adults.

Further, the data augmentation such as 3D rotation, body-shape variation, and noise was applied. Moreover, the random sampling based on the different lengths of data sequence provided the effect of data augmentation. For the NTU RGB+D dataset, the input length ranged from 32–128, and for the ETRI-Activity3D dataset, it ranged from 32–200.

### A. Action recognition on NTU RGB+D and ETRI-Activity3D

For an objective comparison with the other reported networks, FSA-CNN was applied to the NTU RGB+D dataset and our new ETRI-Activity3D dataset. The accuracies of other methods on NTU RGB+D have already been reported in their studies. However, their accuracies on ETRI-Activity3D have been measured in our experiments using their publicized codes.

The NTU RGB+D dataset suggests two experimental protocols: cross-subject and cross-view. We followed the same protocols. The experimental results and comparisons are reported in Table III. As shown in the table, the proposed method achieved the best performance in both protocols.

Additionally, we followed the "cross-subject" protocol for the ETRI-Activity3D dataset. To do so, we divided the entire subjects' data into a training and test sets. The training set consisted of 67 subjects', and the test set consisted of the remaining 33 subjects' data. The subject IDs of the test set are

{3, 6, 9, 12 … 99}, and the remaining are of the training set. As shown in Table III, the proposed network achieved the best accuracy, i.e., 90.6%, on the ETRI-Activity3D. Note that the size of ETRI-Activity3D is approximately twice that of NTU RGB+D.

All the methods presented in Table III use only the skeletal data for recognition. Beyond Joint [17], IndRNN [18], and MANs [19] attempted to use an RNN for action recognition. Because RNN-based methods use sequential data, they are not restricted by a fixed input length. However, they are highly sensitive to noise, and they should have the ability to forget the past data at an appropriate time. Owing to these reasons, RNN-based approaches often show lower accuracies, but they are highly robust to variations in input length.

The architectures of SK-CNN [14], Ensem-NN [16], and HCN [15] are similar to that of the proposed one. These methods typically extract spatial and temporal features and combine them to recognize human actions. However, many of these methods suffer from overfitting and slow operating speeds.

ST-GCN [20] and Motif ST-GCN [21] adopted a GCN as a classifier. The GCN is a CNN having links with each node considering the natural connections of joints in the human body. Thus, the classifiers achieve good performances by considering domain knowledge. However, they have some disadvantages like needs of pre-defined adjacency matrix. Moreover, it is impossible to apply to skeleton data with different structures.

Further, in CNN- and GCN-based methods, the temporal variations are less investigated. Additionally, the RNN-based methods exhibit weaknesses in terms of spatial variations. Therefore, the proposed architecture was designed such that it addresses these issues to achieve a better performance. Our network extracts robust features despite the varying input lengths via global pooling. With the help of global max pooling, the feature extractors of our network focus only on the conspicuous patterns. In case of the global average pooling, the extractors are significantly affected by noise because they extract features from a holistic view, resulting in the loss of salient patterns owing to the averaging with noise. Additionally, the activation network adaptively extracts robust features despite the spatio-temporal variations. Because the proposed network is based on a two-stream CNN approach, it inherits the advantages of its original robustness to spatial noise. In summary, the three components of the proposed network enhance robustness to the temporal, spatio-temporal, and spatial domains. Moreover, our network can adapt to the input itself through the mechanism of the activation network.

### B. Analytical experiments

We applied the proposed network to verify the following issues. First, we divided the ETRI-Activity3D dataset into two: the elderly and adults, and trained the two separate networks using each subset to evaluate the cross-age performance. Second, we evaluated the robustness of the proposed network to temporal variations. Finally, we investigated the possibility of extending our network using the 2D skeleton extracted from RGB videos and the combination of 2D and 3D skeletons.

TABLE IV.  PERFORMANCES OF THE PROPOSED NETWORK WHEN TRAINED ON THREE DOMAINS: THE ELDERLY, ADULTS, AND MIXED.

| Train \ Test | Elderly | Adults |
|---|---|---|
| Elderly | 87.7 | 69.0 |
| Adults | 74.9 | 85.0 |
| Mixed | 84.8 | 82.1 |

*1) Analysis of domain difference*

The ETRI-Activity3D dataset consisted of 50 elderly and 50 young adults. For this cross-age experiment, the training and test sets of the cross-subject protocol were split again to generate 4 sets: training and test sets for the elderly and the corresponding two sets for young adults. The subject IDs of the elderly test set were {3, 6, 9, 12 … 48}, and those of the adult test set were {51, 54, 57 … 99}. Then, the two separate networks were trained using the training sets of the elderly and young adults, respectively, and they were then tested using both the test sets.

Additionally, we trained a network using mixed data. In this case, only half of the mixed training dataset was used so that the number of subjects remains the same as the other two. The subject IDs were {1, 4, 7, 10 … 97}. As shown in Table IV, the network trained using the data of the elderly exhibited significantly higher performance on the corresponding test dataset and vice versa; this indicates that there are noticeable differences between the actions of the two domains. The network trained on the combined dataset exhibited relatively good performance in both domains; this proves the need of a dataset containing the data of the elderly.

*2) Robustness to temporal variations*

Additionally, we analyzed the effect of varying input lengths on performance. Note that no additional training or fine-tuning was performed. We trained the network using the data including various lengths and tested it repeatedly using these data. As shown in Fig. 5, this experiment exhibits the robustness of the proposed architecture against temporal variations. Although the input length varies widely from 32–96 frames, the corresponding accuracy is maintained at higher than 83%, which is comparable to the accuracies of the traditional algorithms using a fixed target input length. In case of the 16-frame input, the accuracy abruptly decreased to 45.26%. The convolution layers extract relatively poor features at their boundaries. The poor result observed at the 16-frame input length is thought to be due to the unreliable results of CNN at boundaries.

### C. Extension of the proposed network

Furthermore, we attempted to enhance the recognition performance by combining the 2D skeletal data obtained using OpenPose [28] from the RGB sequences and 3D skeletons captured using Kinect sensors. The networks with similar architectures were applied because both the data have similar structures. The 2D skeletons were extracted from the RGB video sequences, whereas the 3D skeletons were obtained from the depth information. That is, although they appear similar, they originate from different sources, which can lead to richer information.

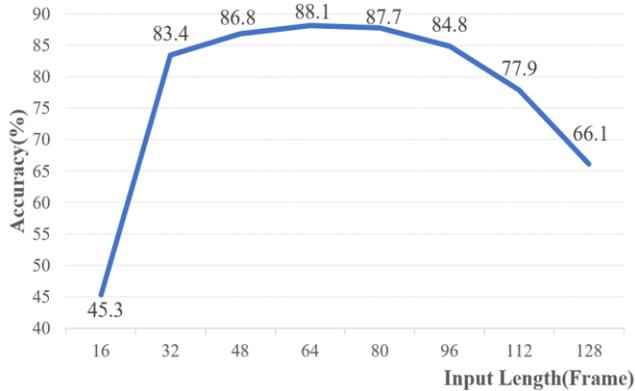

Figure 5. Accuracies of the proposed network including various input lengths.

Table V shows the results of the combination of multimodal methods on NTU RGB+D and ETRI-Activity3D. For comparison, we also experimented with the other state-of-the-art methods that use multimodal information. Although the proposed algorithm with a 2D skeleton only or 3D skeleton only achieved rather competitive results, they were further improved to 91.5% on NTU RGB+D and 93.7% on ETRI-Activity3D when both modalities were combined. The table reveals that our approach achieved second-best accuracy in the NTU RGB+D experiment, and the best accuracy in the ETRI-Activity3D experiment. The combination result exhibited an improvement of 2–3% over those of the single modality-based approaches. This indicates that the two modalities contain complementary information.

Additionally, the table reveals that our approach with the 2D skeleton only achieved quite good results. The 2D-skeleton-based classifier is more convenient in real-world applications because it works without depth cameras, which is an added advantage of the proposed network.

## VI. CONCLUSION

This study proposes a new dataset that includes the actions of the elderly and a novel network for recognizing the actions, and it provides analytical experimental results. ETRI-Activity3D, a new large-scale 3D action recognition dataset containing 112,620 video samples collected using 55 action classes and 100 subjects (50 elderly people and 50 adults), is used. This dataset is expected to be useful for research on action recognition, robotic intelligence, and elderly care. The proposed novel network, FSA-CNN, is evaluated on both NTU RGB+D and the introduced ETRI-Activity3D to prove the advantages. Additionally, the domain differences between actions of the elderly and adults were verified both statistically and experimentally.

To further improve the introduced dataset, an additional 3D activity dataset from the robot's point of view was acquired by visiting the homes of 30 elderly people. This dataset is expected to reflect the real living situation of the elderly better. This additional dataset will be released in the near future for use.

TABLE V. PERFORMANCES OF MULTI-MODAL APPROACHES USING CROSS-SUBJECT PROTOCOL

| Method | Modalities | NTU RGB+D | ETRI-Activity3D |
|---|---|---|---|
| Deep Bilinear Learning [24] | RGBDS | 85.4 | 88.4 |
| Evolution Pose Map [23] | RGB | 78.8 | - |
|  | RGBS | 91.7 | 93.6 |
| c-ConvNet [22] | RGBD | 82.6 | 91.3 |
| **FSA-CNN** | RGB | 87.2 | 90.1 |
|  | S | 88.1 | 90.6 |
|  | RGBS | 91.5 | 93.7 |

ACKNOWLEDGMENT

This work was supported by the ICT R&D program of MSIP/IITP. [2017-0-00162, Development of Human-care Robot Technology for Aging Society]. The protocol and consent of data collection were approved by the Institutional Review Board(IRB) at Suwon Science College.